\documentclass[sigconf, nonacm=true]{acmart} 

\usepackage{algorithm}
\usepackage{algorithmic}
\usepackage{enumitem}
\usepackage{subcaption}
\usepackage{graphicx}
\usepackage{hyperref}
\usepackage{balance}
\usepackage{multirow}

\newcommand{\NA}{---}
\DeclareMathOperator*{\argmax}{argmax}
\newcommand{\ts}{\textsuperscript}

\AtBeginDocument{%
  \providecommand\BibTeX{{%
    \normalfont B\kern-0.5em{\scshape i\kern-0.25em b}\kern-0.8em\TeX}}}

\copyrightyear{2023}
\acmYear{2023}
\setcopyright{cc}%rightsretained
\acmConference[CIKM '23]{Proceedings of the 32nd ACM International Conference on Information and Knowledge Management}{October 21--25, 2023}{Birmingham, United Kingdom}
\acmBooktitle{Proceedings of the 32nd ACM International Conference on Information and Knowledge Management (CIKM '23), October 21--25, 2023, Birmingham, United Kingdom}
\acmDOI{10.1145/3583780.3615266}
\acmISBN{979-8-4007-0124-5/23/10}
\begin{document}
\title{DynED: Dynamic Ensemble Diversification in Data Stream Classification}

\author{%
Soheil Abadifard}
\email{soheil.abadifard@bilkent.edu.tr}
\orcid{0000-0002-2980-4251}
\affiliation{
  \institution{Bilkent University}
  \city{Ankara}
  \country{Turkey}
}
\author{Sepehr Bakhshi}
\email{sepehr.bakhshi@bilkent.edu.tr}
\orcid{0000-0003-2292-6130}
\affiliation{%
  \institution{Bilkent University}
  \city{Ankara}
  \country{Turkey}
  }
\author{Sanaz Gheibuni}
\email{sanaz.gheibuni@bilkent.edu.tr}
\orcid{0009-0005-4349-7568}
\affiliation{%
    \institution{Bilkent University}
    \city{Ankara}
  \country{Turkey}
  }
\author{Fazli Can}
\email{canf@cs.bilkent.edu.tr}
\orcid{0000-0003-0016-4278}
\affiliation{%
  \institution{Bilkent University}
  \city{Ankara}
  \country{Turkey}
}

\renewcommand{\shortauthors}{Soheil Abadifard, Sepehr Bakhshi, Sanaz Gheibuni, and Fazli Can}

\begin{abstract}
  Ensemble methods are commonly used in classification due to their remarkable performance. Achieving high accuracy in a data stream environment is a challenging task considering disruptive changes in the data distribution, also known as concept drift. A greater diversity of ensemble components is known to enhance prediction accuracy in such settings. Despite the diversity of components within an ensemble, not all contribute as expected to its overall performance. This necessitates a method for selecting components that exhibit high performance and diversity. We present a novel ensemble construction and maintenance approach based on MMR (Maximal Marginal Relevance) that dynamically combines the diversity and prediction accuracy of components during the process of structuring an ensemble. The experimental results on both four real and 11 synthetic datasets demonstrate that the proposed approach (DynED) provides a higher average mean accuracy compared to the five state-of-the-art baselines.
\end{abstract}





\maketitle

\section{INTRODUCTION} \label{sec: intro}
The ability to extract meaningful insights out of an endless flow of incoming data is crucial nowadays, and data stream classification methods have made this task more feasible. As more organizations move towards a more technology-driven environment, there has been an alarming increase in generated large amounts of real-time information coming from different sources, such as social media platforms, sensor-based systems, healthcare, etc. Data stream classification techniques offer rapid processing capabilities. This allows analysts to harness valuable insights at lightning speed. As a result, decision-making processes can be carried out promptly to minimize risks while improving performance.

Dealing with the dynamic nature of data streams is one of the main challenges with data stream classification due to changes in the data distribution. This phenomenon is known as concept drift \cite{tsymbal2004problem, WANG2011247, ijcai2022p788, gulcan2023unsupervised, bakhshi2021broad}, and necessitates a learning paradigm capable of handling it. Ensemble approaches combine multiple possibly weak classifiers to improve model performance, robustness, resilience distributivity, and redundancy \cite{vardi2020efficiency}. These approaches have the ability to adapt to the changes in data distributions while maintaining high levels of accuracy \cite{wankhade2020data, wares2019data, bonab2018goowe}.%%Ensemble approaches are techniques that combine multiple weak classifiers to enhance the performance, robustness, and resilience \cite{vardi2020efficiency} of the model. , surpassing that of single classifiers

The discrepancies in predictions provided by individual ensemble components are referred to as diversity. In the ensemble learning setting, maintaining diversity among the individual ensemble components is one of the main challenges. Exposure of the data stream environment to various concept drifts and the fast arrival rate of data items makes this challenge even harder. High-diversity ensembles demonstrate better performance in the presence of concept drift \cite{5156502, minku2011online} even with fewer components \cite{8606435}. The performance of ensemble components can decrease drastically when concept drift occurs. To maintain high accuracy, it is critical to detect the concept drift and update or replace the impacted ensemble components \cite{minku2011online}.

Several approaches to handle the difficulties mentioned above have been proposed. Leveraging Bagging (LevBag) \cite{bifet2010leveraging} combines bagging's simplicity with additional randomization of component inputs and outputs. This randomization can help individual components in an ensemble make different predictions. Concerning diversity, the Adaptive Random Forest (ARF) \cite{gomes2017adaptive} uses a local randomization strategy to retain diversity among ensemble components. This method uses different random selections of features for each component in the ensemble, encouraging diversity among individual components. 

Compared to the prior methods, the Streaming Random Patches (SRP) \cite{gomes2019streaming} combines random subspaces and online bagging to achieve competitive prediction performance. As a result, they indirectly increase the diversity of the ensemble components. Lastly, Kappa Updated Ensemble (KUE) \cite{cano2020kappa} combines online and block-based methods and uses the Kappa statistic to weigh and select classifiers dynamically. To increase the diversity, each base learner is trained with a different subset of features. Additionally, new instances are added to each base learner with a specific probability based on the Poisson distribution.

The challenges of data stream classification and the efforts of the previous solutions to increase diversity within their ensembles have motivated further investigation. The aim is to determine how to add more variety and prune the redundant or ineffective components \cite{10.1007/978-3-030-89814-4_50, elbasi2021onthefly, wozniak2023active} in an ensemble to handle concept drift better, and maintain high accuracy.

The following are the main contributions of this research. We
\begin{itemize}[leftmargin=*,align=left]
    \item Propose a novel ensemble construction and maintenance approach, called DynED (\textbf{Dyn}amic \textbf{E}nsemble \textbf{D}iversification), based on the principles of the Maximal Marginal Relevance (MMR) concept;
    \item Adjust the diversity parameter dynamically to cope with the data stream to have high diversity in case of severe drifts;
    \item Experiment with 15 datasets with varying drift types and compare our results with those of the state-of-the-art methods.
    %\item Examine the influence of the $\lambda$ parameter on ensemble diversity and accuracy.
    %\item Examine the influence of ensemble component count on accuracy. 
\end{itemize}
%Section \ref{sec: prop} provides a comprehensive explanation of DynED. Section \ref{sec: eval} presents and compares the results of experiments with baseline methods. Section \ref{sec: conclusion} concludes the findings and outlines potential future research directions.
%A comprehensive explanation of DynED is provided in Section \ref{sec: prop}. Subsequently, the results of the experiments are presented and compared to those of baseline methods in Section \ref{sec: eval}. Finally, the findings are concluded with a discussion, and potential directions for future research are outlined in Section \ref{sec: conclusion}.
\section{PROPOSED APPROACH} \label{sec: prop}
\begin{figure*}
\centering
  \includegraphics[width=0.72\textwidth]{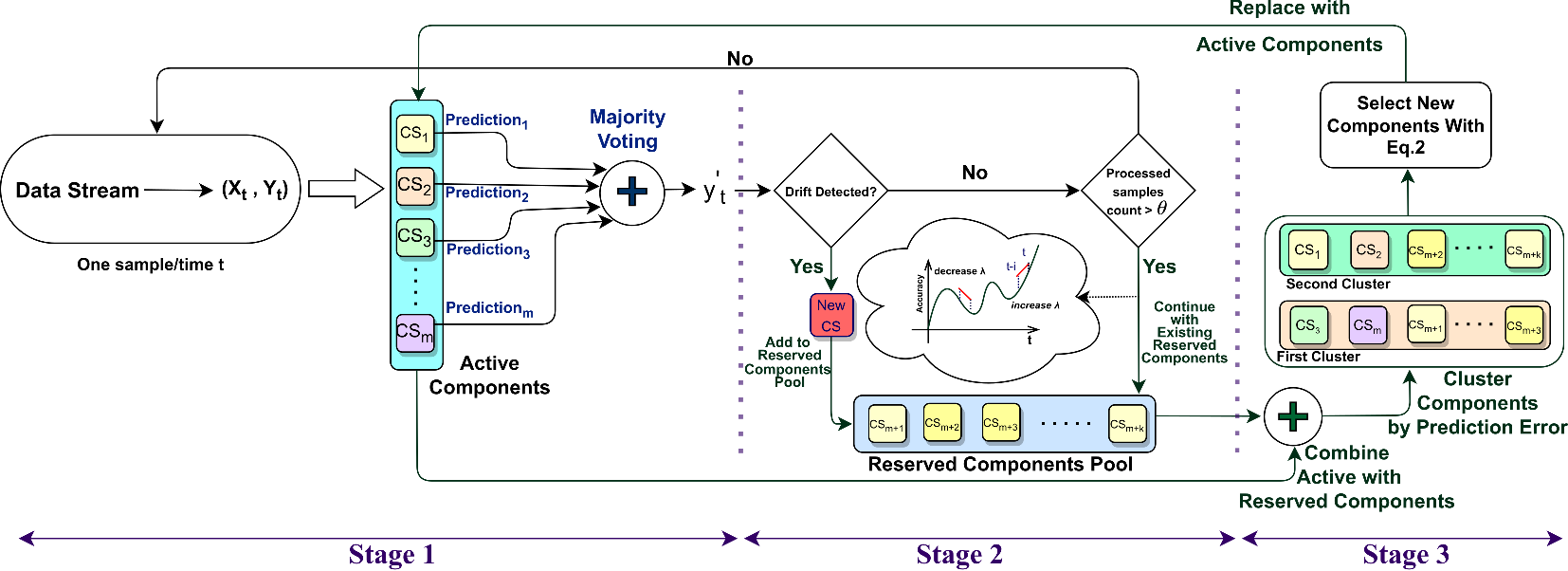}
  \caption{Ensemble construction and maintenance using DynED. Stage 1: Predicting, majority voting, and training. Stage 2: Detecting drifts, adding new components, and updating the diversity parameter. Stage 3: Selecting new components.}
  \label{fig:teaser}
\end{figure*}

\subsection{Using MMR in Data Stream Classification} \label{subsec: mmr}
Maximal Marginal Relevance (MMR) \cite{carbonell1998use} is a diversity-based ranking method that minimizes redundancy while maintaining the relevance of a query in a document set. It is useful for text-document summarization, response extraction \cite{mao2020multi, adams2022combining}, and document re-ranking \cite{carbonell1998use}. Formally, the MMR method follows Eq. \ref{eq:1} to rank the documents:

\begin{equation} \label{eq:1}
    \footnotesize
    \text{MMR} = \argmax_{D_i \in \texttt{R\textbackslash{}S}}[\lambda\times\text{sim}(D_i, q) - (1-\lambda)\times\max_{D_j \in S} \text{sim}(D_i, D_j)]
\end{equation}

In Eq.\ref{eq:1}, $D_i$ represents a document, $R$ is a ranked list of the documents, $S$ is a set of the selected documents, $\lambda$ is a parameter that balances accuracy and redundancy, and $sim(D_i,q)$ measures the relevance between document $D_i$ and query $q$. When $\lambda = 1$, MMR calculates the relevance-ranked list; when $\lambda=0$, it calculates a ranking that maximizes diversity among the documents in $R$. MMR optimizes a linear combination of relevance and diversity criteria for values of $\lambda$ between 0 and 1.
\begin{table}[ht]
  \centering
  \tiny
  \caption{Symbols \& default hyperparameter values of DynED}
  %\description{This table illustrates the Symbols, Notations, and initial values used in the proposed method.}
  \label{tab:Symbols}
  \begin{tabular}{l l c}
    \toprule
    Symbol & Meaning & Default value\\
    \midrule
    %$\xi$ & Pool of ensemble components $\xi=\{CS_1,..., CS_m\}$ & \NA\\
    $\xi_{slc}$ & Set of selected components & \NA\\
    $I$ & Intensity of changes in the accuracy & \NA\\
    %clustering & K-means & \NA \\
    %DD & Drift detector (ADWIN) & 7e-1 \\
    $\theta$ & Threshold for the count of processed samples & 100\\
    $\lambda$& Controlling parameter of similarity and accuracy & 0.6\\
    $ \Delta_{\lambda}$ & Variation in $\lambda$ parameter & 0.1\\
    $win_s$& Window size & 500\\
    $init_c$& Initial components count & 5\\
    $init_s$& Initial training sample size per component & 50\\
    $pool_s$& Maximum size of the component pool & 500\\
    $add_s$& \# of components to add in each step & 5\\
    $cls_s$& \# of components to select from each cluster& 10\\
    $slc_s$ & \# of active components for classification & 10\\
    $error_s$& Prediction list size to calculate similarity on & 50\\
    $cls_{slc}$& Set of selected components based on clustering & \NA \\
    $cls_{rst}$& Set of not selected components based on clustering & \NA\\
  \bottomrule
\end{tabular}
\end{table}
It is necessary to make changes in its definition in order to adapt the MMR method for selecting and ranking ensemble components. In terms of ensemble components, the first part of Eq. \ref{eq:1}, which calculates the relevance of $D_i$ to query $q$, is replaced with the accuracy of each component. It is represented as \textit{"$\lambda\times\text{acc}(CS_i, Ins)$"}, where $\xi=\{CS_1, CS_2,..., CS_m\}$ is the set of ensemble classifiers, $CS_i$ represents each component, and $Ins$ are the previously seen instances of the data stream.

The second part of the Eq. \ref{eq:1} determines a pairwise similarity between the documents. However, evaluating the diversity of ensemble components is difficult since there is no commonly agreed-upon formal definition of diversity. Several methods are available for determining the pairwise diversity of classifiers in terms of correct/incorrect (oracle) outputs, such as \textit{Correlation Coefficient P (CP)}, \textit{double-fault measure (DF)}, \textit{Disagreement Measure (DM)}, and \textit{$Q$ statistic} \cite{tsymbal2005diversity, kuncheva2001ten, kuncheva2003measures}. To adapt the second part of the Eq. \ref{eq:1} to the context of component diversity, we replace it with the DF diversity measuring method (the procedure of diversity measure selection is explained in section \ref{subsec: setup}). Therefore, the second part of the formal equation turns into \textit{"$(1-\lambda)\times\max_{CS_j\in S}sim(CS_i, CS_j)$"}. The final version of the MMR method for our task is presented in Eq. \ref{eq:3}:

\begin{equation} \label{eq:3}
  \footnotesize
  \centering
  \text{MMR}=\argmax_{CS_i\in\texttt{R\textbackslash{}S}}[\lambda\times\text{acc}(CS_i, Ins) - (1-\lambda)\times\max_{CS_j\in S} sim(CS_i, CS_j)]
\end{equation}

%Diversity measures calculate diversity between components by default, and the MMR method is based on similarity, derived by subtracting diversity from unity.
The MMR method utilizes a measure of similarity, which can be derived from the complement of a diversity measure.
\subsection{Dynamic Ensemble Diversification: DynED} \label{subsec: dyned}
\textit{The working principles of our approach: }DynED aims to dynamically ensure high accuracy by increasing diversity in the presence of concept drift and otherwise by decreasing it to reduce exposure to underperforming ensemble components. The pseudo-code for DynED is provided in Algorithms \ref{alg: main} and \ref{alg: DynED}.
\begin{algorithm}[b]
    \caption{DynED: Main flow.}
    \centering
    \tiny
    \label{alg: main}
    \begin{algorithmic}[1]
        \REQUIRE $D$: data stream, $W$: sliding window, $DD$: drift detector, $init_s$, $init_c$, $win_s$, $pool_s$, $error_s$, $add_s$, $ \Delta_{\lambda}$
        \ENSURE $\hat{y_k}$: prediction of each sample at time $t$
        \STATE $\xi, \xi_{slc} \leftarrow \text{initialize } init_c \text{ components and train on } init_s \text{ samples}$
        \WHILE{$D$ has more samples}
            \STATE $\hat{y_k} \leftarrow \text{use selected components }\xi_{slc}\text{to predict using majority voting} $
            \STATE $W \leftarrow$ add sample to the sliding window
            \STATE $DD\leftarrow \text{update drift detector using correct/incorrect predictions}$
            \STATE train $\xi_{slc}$ on sample
            \IF{$DD$ detects drift}
                \STATE $\xi \leftarrow$ generate and train $add_s$ new component on data in $W$, add to the pool
            \ENDIF
            \IF{$\text{passed samples count} \ge \theta$ or new component is added}
                \STATE $I \leftarrow$ calculate the intensity of accuracy change to update the $\lambda$ parameter dynamically
                \IF{$I \geq 0$}
                    \STATE increase $\lambda$ by $ \Delta_{\lambda}$ until it reaches 1
                \ELSE
                    \STATE decrease $\lambda$ by $ \Delta_{\lambda}$ until it reaches 0
                \ENDIF
                \STATE $\xi_{slc}$, $\xi\leftarrow$ update $\xi_{slc}$ and $\xi$ using Algorithm \ref{alg: DynED}
            \ENDIF
        \ENDWHILE
    \end{algorithmic}
\end{algorithm}
\newline\textit{Stage 1:} The primary process includes predicting new samples and training selected components, which is outlined in Algorithm \ref{alg: main}. As this method operates online, line 3 of Algorithm \ref{alg: main} uses selected components to predict new samples using Majority Voting \cite{8001122, 8907028}.
\newline\textit{Stage 2:} The Drift Detector, ADWIN \cite{bifet2007learning}, is updated using the correct/incorrect predictions. If drift is detected, a new classifier is generated and trained on the last seen data available in the sliding window and then added to the reserved component pool $\xi$ (lines 5-9 of Algorithm \ref{alg: main}). Suppose a new component is added or the processed sample count passes the $\theta$ threshold. In that case, the algorithm updates the $\lambda$ parameter to reflect the proper value of diversity based on the intensity of accuracy changes (lines 10-16). Then Algorithm \ref{alg: DynED} is called to update $\xi_{slc}$ and $\xi$. The intensity of changes in the accuracy is computed using the formula "$acc(t) - acc(t-i) \mathbin{/} i$" where $acc(t)$ denotes the accuracy of the ensemble model at the t\ts{th} sample and $i$ is equivalent to $\theta$ in DynED.
\begin{algorithm}
    \caption{DynED: Component selection.}
    \tiny
    \centering
    \label{alg: DynED}
    \begin{algorithmic}[1]
        \REQUIRE $W$: data window, $\xi$, $\xi_{slc}$, $slc_s$, $\lambda$, $error_s$, $pool_s$, $cls_s$
        \ENSURE $\xi$, $\xi_{slc}$
        \STATE $P \leftarrow \text{ combine }\xi \text{ and }\xi_{slc}$ sorted by accuracy
        \WHILE{$len(p) \geq pool_s$}
            \STATE remove the last component
        \ENDWHILE
        \STATE $PE\leftarrow$ obtain prediction error for all components recalling data from $W$
        \STATE $cls\leftarrow$ cluster components based on $PE$ "(two clusters)"
        \STATE $cls_{slc}\leftarrow$ select $cls_s$ components from each cluster based on the accuracy
        \STATE $cls_{rst}\leftarrow$ store rest of components
        \STATE $\xi_{slc}$, $\xi_{rst} \leftarrow$ apply Eq. \ref{eq:3} on $cls_{slc}$ to obtain a new set of components to continue prediction %MMR($cls_{slc}$, $Size_{slc}$)
        \STATE $\xi\leftarrow cls_{rst} + \xi_{rst}$
        \RETURN $\xi$, $\xi_{slc}$
    \end{algorithmic}
\end{algorithm}
\newline\textit{Stage 3:} Algorithm \ref{alg: DynED} selects a diverse set of components using an adapted MMR method presented in Eq. \ref{eq:3}.  The algorithm combines previously selected components with those in the reserve pool and maintains a fixed component count by sorting them based on accuracy and removing poorly performing components. In line 5 of Algorithm \ref{alg: DynED}, prediction errors for all components on previous samples held in sliding window are obtained. In the following line, components are clustered using the K-means clustering algorithm \cite{hartigan1979algorithm} into two groups based on these prediction errors to apply the selection method effectively. After clustering, $cls_s$ high-performance components are selected from each cluster, resulting in a total of $2\times cls_s$ components out of $N$ where $N\leq pool_s$. In line 9 of Algorithm \ref{alg: DynED}, an adapted MMR method is applied to perform the final selection step. This algorithm outputs a new set of selected components to predict new incoming samples of the stream actively and updates the reserved components pool as a result. The way an ensemble structure is constructed and maintained by DynED is illustrated in Figure \ref{fig:teaser}.

\renewcommand{\thefigure}{2}
\begin{figure*}
\centering
    \begin{minipage}{0.36\textwidth}
    \begin{subfigure}{0.495\textwidth}
    \includegraphics[width=\linewidth]{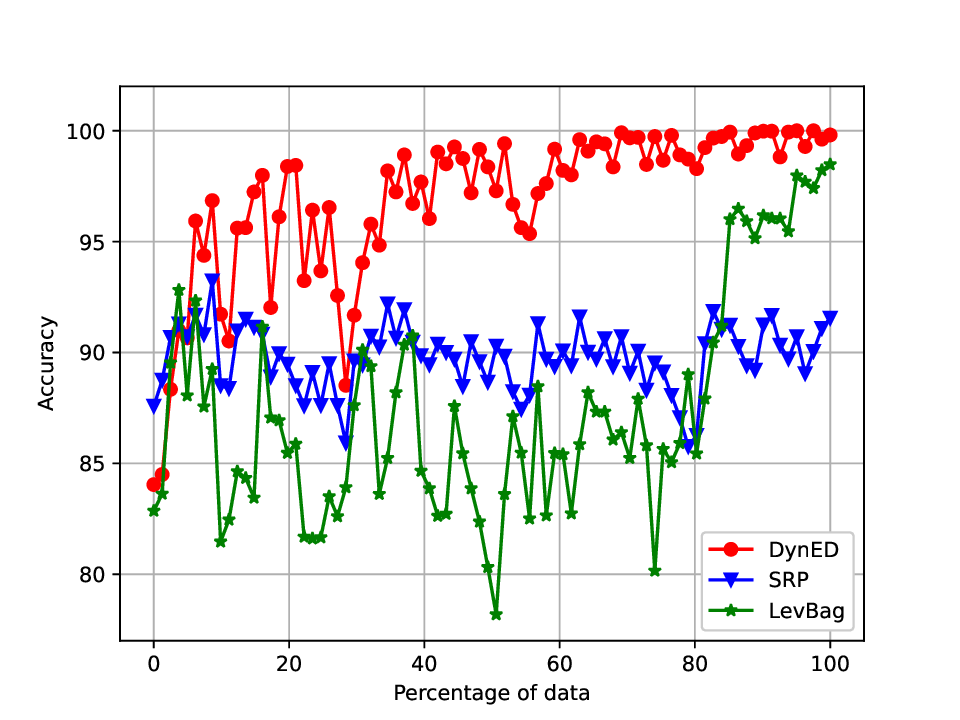}
    \subcaption{Poker.}
    \label{fig: Poker}
    \end{subfigure}
    \begin{subfigure}{0.495\textwidth}
    \includegraphics[width=\linewidth]{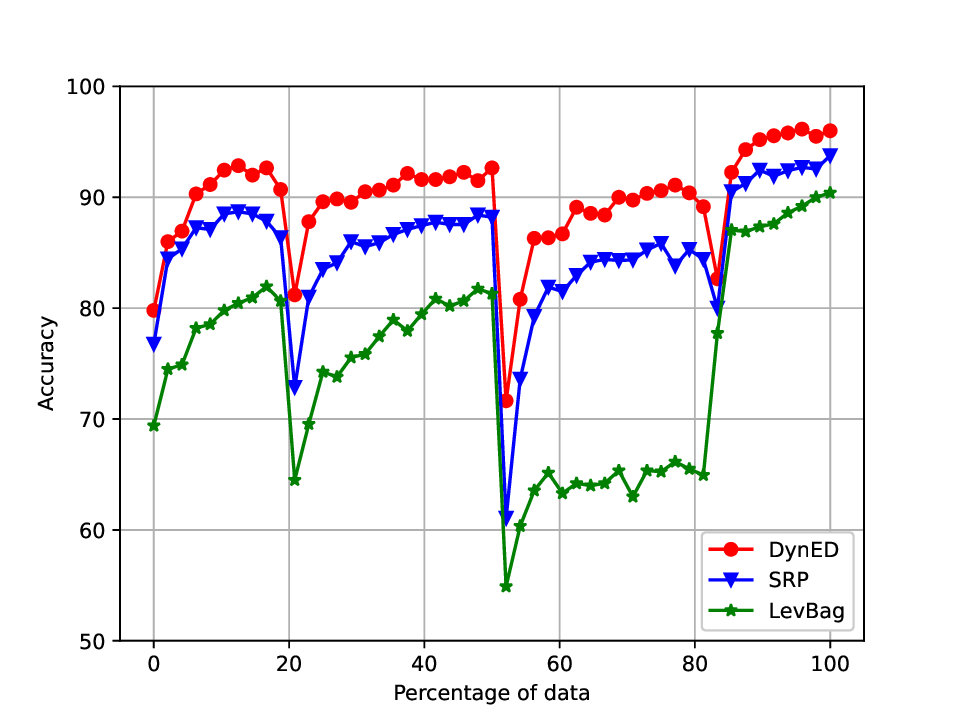}
    \subcaption{Agrawal-4567.}
    \label{fig: Agrawal-4567}
    \end{subfigure}
    \begin{subfigure}{1\textwidth}
    \centering
    \includegraphics[width=\linewidth]{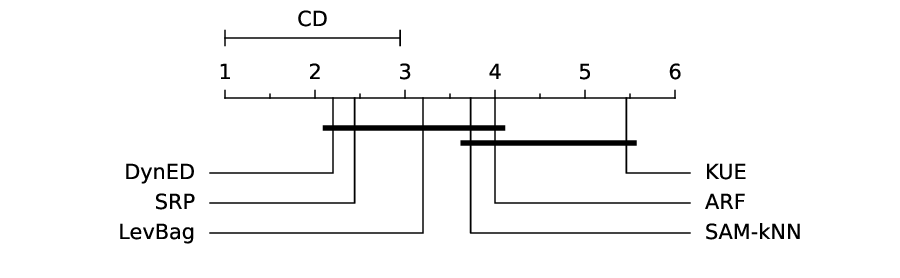}
    \subcaption{Critical distance diagram.}
    \label{fig: cd}
    \end{subfigure}
    \captionof{figure}{(a, b): Prequential temporal accuracy results of DynED, and (c): Critical distance diagram, CD= 1.946.}
    \label{fig: diag}
    \end{minipage}
  \begin{minipage}{0.58\textwidth}
    \centering
    \captionof{table}{Characteristics of the datasets, average interleaved-test-then-train accuracy, and the rankings of the methods for each dataset. DT: Drift Type, |X|: No. of features, |y|: No. of classes, |$\mathcal{D}$|: No. of Samples}
    \resizebox{1\textwidth}{!}{
    \begin{tabular}{c|l|crrr|rrrrrr}
\multicolumn{1}{l|}{}                                     & \textbf{Name}&\textbf{DT}&\textbf{|X|}&\textbf{|y|}&\textbf{|$\mathcal{D}$|}&DynED     & KUE            & SRP            & ARF           & SAM-kNN          & LevBag          \\ \hline
\multirow{5}{*}{\rotatebox[origin=c]{90}{\textbf{Real}}}  & Electricity \cite{harries1999splice}   & U    & 6   & 2   & 45,312   & \textbf{87.88}  & 76.77          & 87.85          & 85.29         & 79.16            & 87.28           \\
                                                          & Poker \cite{Dua:2019}                  & U    & 10  & 10  & 829,201  & \textbf{96.96}  & 81.68          & 89.74          & 80.65         & 82.41            & 85.99           \\
                                                          & Cover Types                            & U    & 54  & 7   & 581,012  & \textbf{94.66}  & 84.30          & 94.07          & 90.06         & 93.87            & 90.94           \\
                                                          % Airlines                               & U    & 7   & 2   & 539,383   \\
                                                          & Weather \cite{elwell2011incremental}   & U    & 8   & 2   & 18,152   & 78.17           & 72.74          & \textbf{78.79} & 78.28         & 78.24            & 77.87           \\ \hline
\multirow{10}{*}{\rotatebox[origin=c]{90}{\textbf{Synthetic}}} & Interchanging RBF \cite{7837853}  & A    & 2   & 15  & 200,000  & 93.70           & 71.77          & \textbf{97.44} & 97.15         & 94.19            & 95.34           \\
                                                          & LED-drift                              & A    & 7   & 10  & 100,000  & \textbf{99.99}  & 95.33          & 99.32          & 96.65         & 99.94            & 99.96           \\
                                                          & MG2C2D \cite{dyer2013compose}          & I \& G & 2   & 2   & 200,000  & 92.68           & 93.17          & 94.53          & 94.49         & 94.52            & \textbf{94.54}  \\
                                                          & Moving Squares \cite{7837853}          & I    & 2   & 4   & 200,000  & 85.02           & 30.05          & 86.95          & 56.95         & \textbf{97.34}   & 87.77           \\
                                                          & Rotating Hyperplane \cite{7837853}     & I    & 10  & 2   & 200,000  & 84.69           & 83.36          & 85.79          & 81.53         & \textbf{86.25}   & 86.05           \\
                                                          & SEA-Abrupt-012                         & A \& R & 3   & 2   & 100,000  & \textbf{90.61}  & 86.80          & 90.16          & 90.20         & 88.61            & 89.90           \\
                                                          & SEA-Abrupt-123                         & A \& R & 3   & 2   & 100,000  & \textbf{90.10}  & 86.60          & 89.79          & 89.61         & 88.43            & 89.60           \\
                                                          & SEA-Gradual-012                        & G \& R & 3   & 2   & 100,000  & \textbf{80.33}  & 77.28          & 79.83          & 79.77         & 77.20            & 79.76           \\
                                                          & SEA-Gradual-123                        & G \& R & 3   & 2   & 100,000  & \textbf{80.30}  & 77.56          & 79.82          & 79.57         & 76.93            & 79.60           \\
                                                          & Agrawal-4567                           & G \& I & 9   & 2   & 100,000  & \textbf{89.95}  & 75.41          & 84.29          & 73.64         & 73.91            & 74.77           \\
                                                          & Mixed-12                               & A \& R & 4   & 2   & 100,000  & \textbf{98.47}  & 89.45          & 93.44          & 97.02         & 98.20            & 94.71           \\ \hline
                                                          & Average Mean                           & \NA  & \NA & \NA & \NA      & \textbf{89.57}  & 78.82          & 88.78          & 84.72         & 87.28            & 87.60           \\ \hline
                                                          & Rank                                   & \NA  & \NA & \NA & \NA      & \textbf{2.20}   & 5.46           & 2.44           & 4.00          & 3.73             & 3.20            \\
                                                          %& Waveform                              & -    & 21  & 3    & 100,000  
    \end{tabular}
    \label{tab: data}
    }
    \end{minipage}
\end{figure*}

\subsection{Time Complexity Analysis of Component Selection} \label{subsec: time}
The time complexity of Algorithm \ref{alg: DynED}, which employs the reformulated MMR method, is as follows. In line 5 of Algorithm \ref{alg: DynED}, the prediction errors of all classifier components are obtained, the time complexity is $\mathcal{O}(p)$, where $p$ represents the classifier component count and $p \leq pool_s$. Lines 6, 7, and 8 of Algorithm \ref{alg: DynED} involve clustering the classifier components based on their prediction errors and selecting $cls_s$ from each cluster, where $cls_s < p$. The time complexity of these operations is $\mathcal{O}(k \times p \times itr)$, where $itr$ is the number of iterations in the clustering process, and $k$ is the number of clusters. $itr$ and $k$ are considered as constant as they are not hyperparameters for DynED. In line 10, which applies the reformulated MMR method, the time complexity can be broken down as follows: calculating the pairwise similarity of classifier components using any diversity measure has a time complexity of $\mathcal{O}(n^2)$, where $n$ represents the number of classifier components extracted by the clustering step ($2\times cls_s$). Applying the reformulated MMR method itself has a time complexity of $\mathcal{O}(n \times j)$, where $j = slc_s$. Therefore, the overall time complexity of Algorithm \ref{alg: DynED} is $\mathcal{O}(p + k \times p \times itr + n \times j + n^2)$. The dominant term in this time complexity analysis is $n^2$. Hence, the algorithm’s time complexity can be approximated as $\mathcal{O}(n^2)$.
% the term ($n^2$) dominates the rest. Thus, for the sake of simplicity, $\mathcal{O}(n^2)$ can be assumed as the overall time complexity of Algorithm \ref{alg: DynED}.

\section{EXPERIMENTAL EVALUATION} \label{sec: eval}

\subsection{Datasets} \label{subsec: data}
To assess the performance of our model, we conduct experiments using 15 datasets (Four real and 11 synthetic datasets) and compare them to the baseline models. The datasets cover a wide range of concept drift scenarios. Our experiments include all four types of drift: Gradual (G), Incremental (I), Abrupt (A), Recurring (R), and (U) stands for Unknown drift type. The synthetic datasets based on LED, SEA, Agrawal, and Mixed generators are created using the scikit-multiflow library \cite{montiel2018scikit} and MOA framework \cite{DBLP:journals/jmlr/BifetHKP10}. The LED dataset has seven drifting features without noise. The Agrawal dataset uses four classification functions, and the SEA dataset uses three classification functions to synthesize drift. The description of the datasets is shown in Table \ref{tab: data}.

\subsection{Setup} \label{subsec: setup}
In our study, we evaluate four diversity measures: \textit{Correlation Coefficient P (CP)}, \textit{Double-Fault measure (DF)}, \textit{Disagreement Measure (DM)}, and \textit{$Q$ statistic}. We apply each measure to Eq. \ref{eq:3} across all datasets to determine the most suitable diversity measure for DynED. Our results show that the DF has the highest average mean accuracy compared to that of CP, DM, and $Q$-statistic with respective average mean accuracies of 89.57, 88.14, 89.43, and 89.38. Therefore, we choose DF as the diversity measure in DynED. 

We evaluate the performance of DynED against five state-of-the-art baselines including LevBag \cite{bifet2010leveraging}, SAM-kNN \cite{losing2016knn}, ARF \cite{gomes2017adaptive}, and SRP \cite{gomes2019streaming}. These baselines are assessed using the Massive Online Analysis (MOA) \cite{DBLP:journals/jmlr/BifetHKP10} framework with default hyperparameters. For KUE \cite{cano2020kappa}, we use the source code available on their GitHub for evaluation \footnote{\url{https://github.com/canoalberto/imbalanced-streams}}. DynED is implemented in Python 3.8 using the scikit-multiflow \cite{montiel2018scikit} library, with a Hoeffding Tree as the base classifier, and \textit{split-confidence} set to 9e-1 and \textit{grace-period} set to 50. All baseline models were evaluated using the Interleaved-test-then-train approach. The codes and datasets for experiments are publicly available, and all experiments and results are reproducible \footnote{\url{https://github.com/soheilabadifard/DynED}}.

The selection of appropriate hyperparameters is a critical aspect of all machine learning methods, including DynED. After conducting tests with various hyperparameters based on the grid search method, we determined the selected values presented in Table \ref{tab:Symbols}. These values serve as the default hyperparameters in DynED and are not tailored to any specific dataset. It should be noted that the first 250 samples of each dataset are used as a warm-up, and they are not involved in the accuracy calculations and final results presented in Table \ref{tab: data}.
%\begin{figure}[ht]
%\centering
%  \includegraphics[width=0.45\textwidth]{test-dunn.eps}
%  \caption{Critical distance diagram for overall accuracy.}
%  \label{fig: cd}
%\end{figure}

\subsection{Results and Discussion} \label{subsec: results}
The overall accuracy of each method applied to each dataset is presented in Table \ref{tab: data}, with the highest scores emphasized in bold. A comparative analysis reveals that DynED outperforms the baselines in 10 out of 15 datasets, particularly in three out of four real and seven out of 11 synthetic datasets. Furthermore, when the average mean accuracies are ranked in descending order, DynED emerges as the top performer with an average rank of 2.20. A closer examination of Table \ref{tab: data} and Figure \ref{fig: diag}.a and Figure \ref{fig: diag}.b indicates that DynED provides robustness in the case of gradual and recurrent drift types, outperforming the baselines in nearly all datasets that exhibit these drift types. However, DynED’s performance declines in the presence of incremental drift, struggling to maintain high accuracy levels throughout the stream. Nonetheless, when confronted with abrupt drifts, DynED effectively captures and addresses the drift by employing Eq. \ref{eq:3} to increase diversity among components, resulting in enhanced performance as evidenced by the plots in Figure \ref{fig: diag}. Overall, the results suggest that DynED is a promising method for the online classification of data streams.
To assess the statistical significance of the employed methods, the \textit{Friedman Test} is utilized in conjunction with the \textit{Nemenyi post-hoc analysis} \cite{demvsar2006statistical}. We calculate the \textit{Critical Distance} as CD= 1.946. In our experiment, $\alpha=0.05$. The statistical test analysis, as presented in Figure \ref{fig: diag}.c, reveals that DynED statistically significantly outperforms KUE and achieves a better ranking among the other baseline methods.

%\begin{figure}[ht]
%     \centering
%     \begin{subfigure}[b]{0.236\textwidth}
%         \centering
%         \includegraphics[width=\textwidth]{poker.eps}
%         \caption{Poker}
%         \label{fig: Poker}
%     \end{subfigure}
%     \hfill
%     \begin{subfigure}[b]{0.236\textwidth}
%         \centering
%         \includegraphics[width=\textwidth]{Agrawal.eps}
%         \caption{Agrawal-4567}
%         \label{fig: Agrawal-4567}
%     \end{subfigure}
%        \caption{Prequential temporal accuracy results of DynED and two ranked methods.}
%        \label{fig: diag}
%\end{figure}

\section{CONCLUSION AND FUTURE WORK} \label{sec: conclusion}
This paper presents DynED, a novel ensemble construction and maintenance method that combines diversity and prediction accuracy for data stream classification tasks. It aims to increase the diversity among components in the presence of concept drift in a data stream in order to handle drifts better. The results show that DynED has higher average mean accuracy compared to the baseline models.
In real-world scenarios, data stream environments often face the issue of label scarcity. As a part of future work, we aim to enhance our study for the semi-supervised classification of data streams.
\section{ACKNOWLEDGMENTS}
This study is partially supported by TÜBİTAK grant no. 122E271.
\newpage

\bibliographystyle{ACM-Reference-Format}
\balance
\bibliography{sample-base}
\end{document}